\documentclass{article}
\usepackage{ijcai18}

\usepackage{times}
\usepackage{xcolor}
\usepackage{soul}
\usepackage[utf8]{inputenc}
\usepackage[small]{caption}
\usepackage{amsmath}

\newcommand{\mypar}[1]{\paragraph{#1}}

\title{Boosting Combinatorial Problem Modeling with Machine Learning}

\author{
  Michele Lombardi {\rm and}
Michela Milano
\\
DISI-University of Bologna \\
michele.lombardi@unibo.it,
michela.milano@unibo.it}

\begin{document}

\maketitle

\begin{abstract}
  In the past few years, the area of Machine Learning (ML) has witnessed tremendous advancements, becoming a pervasive technology in a wide range of applications. One area that can significantly benefit from the use of ML is Combinatorial Optimization. The three pillars of constraint satisfaction and optimization problem solving, i.e., modeling, search, and optimization, can exploit ML techniques to boost their accuracy, efficiency and effectiveness. In this survey we focus on the modeling component, whose effectiveness is crucial for solving the problem. The modeling activity has been traditionally shaped by optimization and domain experts, interacting to provide realistic results. Machine Learning techniques can tremendously ease the process, and exploit the available data to either create models or refine expert-designed ones. In this survey we cover approaches that have been recently proposed to enhance the modeling process by learning either single constraints, objective functions, or the whole model. We highlight common themes to multiple approaches and draw connections with related fields of research.
\end{abstract}

\section{Introduction}
Inferring models from observations and studying their properties is what science is about. Models can be descriptive or interpretative, thus enabling the understanding of a system/process behaviour. Models can be predictive, thus describing the future evolution of the system/process dynamic. Models can be prescriptive, thus providing decision support and assessing the effects of decisions on the system/process evolution in the medium or long term. All models are just approximations of the system/process they represent, but some of them are useful as they provide insights to describe, forecast and decide on the system. Their accuracy is essential for the task they have been built for.

In this paper \emph{we focus on the specific case of combinatorial optimization models}, exhibiting discrete decision variables, a combinatorial structure defined by constraints and an objective function showing the direction for improving the solution quality. These models have been traditionally designed via a close interaction between optimization and domain experts, the former encoding the knowledge of the latter in mathematical or symbolic expressions. Combinatorial model design is indeed an iterative process, where subsequent model versions are evaluated by domain experts; these experts assess the model effectiveness and efficiency, and possibly provide feedback to the optimization expert for further refinement. Uncertainty can be taken into account or not, eventually resulting in either stochastic or deterministic models.

Nowadays, in many application domains we have access to data of unprecedented scale and accuracy about the system/process we want to model. Machine Learning (ML) provides techniques to exploit available data and extract value and useful information, which can be employed for modeling and solving combinatorial problems. The field has been widely explored in the last decade, as witnessed by the increasing number of researchers working in using ML in optimization. A good survey can be found in the edited collection \cite{CP-DM2016} on the intersection of Constraint Programming with Machine Learning and Data Mining.


In this paper \emph{we provide a survey on methods for boosting the combinatorial modeling activity through ML methods and algorithms}. ML techniques can be used to learn either constraints that define the combinatorial structure of the problem, or objective functions describing optimization criteria. In this way, not only we learn part of the model from data, but we also have information about the model accuracy, which is otherwise inaccessible if the model is defined through the interaction with domain experts. We review the recent literature on the use of ML to improve the model or learn it when a declarative form is hard to shape even by domain experts.

The paper is organized as follows: Section~\ref{learningConstraints} contains the bulk of the content and deals with approaches that employ ML to obtain (part of) a combinatorial model. In Section~\ref{sec:learn_and_solve} we present approaches relying on an internal approximate models and active learning to directly solve a problem. In Section \ref{relatedAreas} we draw connections with some related areas, most notably black-box optimization (based on surrogate models) and system identification in Control Theory. In Section \ref{Transversal_issues} we describe some horizontal topics common to the discussed approaches that represent important open research directions.

Finally, we open up a bit the scope of the paper by showing in section \ref{general} a picture of the many directions for the integration of ML in combinatorial optimization: this includes search and optimization, portfolio selection and algorithm configuration. Due to space limitations, we do not discuss the (many and interesting) techniques that explore the other direction of integration (i.e. optimization techniques to improve ML algorithms). Concluding remarks are provided in Section~\ref{sec:conclusions}.

\section{Learning Model Components}
\label{learningConstraints}

In this section we consider approaches that use implicit information from a set of examples to obtain part of a combinatorial model, in particular a constraint or an objective function. There are two main approaches to achieve this result: the first (covered in Section~\ref{sec:constr_acq}) aims at extracting information using the \emph{native constraint language} of the solver; the second (in Section~\ref{sec:eml}) consists in \emph{embedding a fully-fledged Machine Learning model} in a combinatorial approach. As an extreme case, in Section~\ref{learningWholeModel} we consider approaches where the ML model makes up (almost) all of the combinatorial model.


\subsection{Learning via a Native Constraint Language}
\label{sec:constr_acq}

Traditionally, combinatorial optimization models are born from iterative interactions between a domain and an optimization expert. Intuitively, ML could support modeling activities by retaining the same process, but replacing the optimization expert with a \emph{constraint acquisition algorithm}, and the domain expert with an example generator. Formally, the approaches in this section aim at learning a model in the form:
\begin{align*}
  \min z & = x_0 & \mathbf{(P1)} \\
  \text{subject to: } & \pi_i(\vec{x}) & \forall i \in I \\
  & \vec{x} \in D_{\vec{x}}
\end{align*}
were $\vec{x}$ is the vector of problem variables and $D_{\vec{x}}$ their domain. The set $I$ contains the indices of all problem constraints, represented here as predicates $\pi_i(\vec{x})$ that must hold in any feasible solution. The $x_0$ variable represents by convention the cost to be minimized, and is absent in pure constraint satisfaction problems. Crucially, \emph{the predicates are defined using the building blocks from the hosting approach} (e.g. global constraints in Constraint Programming, linear equalities or inequalities in Mixed Integer Linear Programming): what changes is just the way they are discovered.

The example generator may be a human, a collection of data, or an existing automated system. In latter case, constraint acquisition can also be considered as a mean to explain in declarative terms the behavior of a procedural or sub-symbolic decision support system. All approaches in this section focus on learning constraints, rather than objective functions: of course nothing prevents a constraint from representing the definition of the $x_0$ variable (as done in P1).

This is the idea behind systems such as \textsc{Conacq} \cite{dblp:journals/ai/bessiereklo17} (in its various versions), \textsc{QuAcq} \cite{dblp:conf/ijcai/bessierechklnqw13}, and model seeker \cite{dblp:conf/cp/beldiceanus12}, which build over the Constraint Programming paradigm, and behind the method in \cite{dblp:conf/ictai/lallouetlmv10}, based on Inductive Logic Programming. Both \textsc{Conacq} and \textsc{QuAcq} operate by picking constraints from a set of potential (instantiated) candidates (called a \emph{bias}) and adding them to a target constraint network.  Model seeker attempts to match (possibly) transformed subsets of variables in the training examples against a collection of (non-instantiated) global constraints; constraints that are compatible with all examples are added to the current model, and a series of simplification steps attempts to remove redundant relations. Since model seeker relies does not need to consider explicitly all possible instantiations in its candidate pool, it can usually deal with a large variety of constraints. The downside is that finding a matching becomes more complicated and requires the use of a heuristic step. The method from \cite{dblp:conf/ictai/lallouetlmv10} attempts instead to learn local rules that are partially independent on the specific values and variables appearing in the examples.

A first major design choice in all such approaches concerns the use of passive or active learning. Methods based on passive learning (e.g. Model seeker, the original \textsc{Conacq}, and the one from \cite{dblp:conf/ictai/lallouetlmv10}) operate on a fixed collection of examples. Conversely, approaches based active learning (e.g. \textsc{QuAcq} and \textsc{Conacq.2}) generate candidate examples themselves and query the generator (which in this case is instead a constraint checker) for their validity. Active learning enables convergence using a smaller number of examples, but is not applicable when a constraint checker is not available (e.g. when working on collections of historical data).

Among the mentioned approaches, only model seeker can work using just positive examples, which makes it well suited to deal with historical data. The system can also be used to obtain, based on a handful of examples, a candidate list of global constraints for modeling the problem. \textsc{Conacq} and \textsc{QuAcq} employ both positive and negative examples (which may be a disadvantage), but are capable of using negative examples to quickly rule out large sets of constraints from the bias (which is a considerable advantage). \textsc{QuAcq} has the peculiarity of relying on \emph{partial} examples, where only some of the problem variables are instantiated. This allows to speed up convergence both from a theoretical and practical perspective, giving the algorithm its namesake (QUick ACQuisition).

Finally, the method from \cite{dblp:conf/ictai/lallouetlmv10} learns a model using an intermediate representation; this is loosely inspired by modeling languages such as AMPL, OPL or MiniZinc, which make a clear distinction between the problem structure and its parameters. Thanks to this design choice, the approach is able to learn parameter-free models, and to generalize results obtained on smaller instances to larger ones. The price to pay for this impressive feat is a more complex formalism and a somewhat reduced expressivity.

\subsection{Incorporating Machine Learning Models}
\label{sec:eml}

Unlike the approaches described in Section~\ref{sec:constr_acq}, the methods considered here attempt to incorporate a fully-fledged Machine Learning model within a combinatorial optimization model. Formally, these works deal with problems in the form:
\begin{align*}
  \min z & = x_0 & \mathbf{(P2)} \\
  \text{subject to: } & \pi_i(\vec{x}) & \forall i \in I \\
  & \nu_m(\vec{x}_{m,in}, \vec{x}_{m,out}) & \forall m \in M \\
  & \vec{x} \in D_{\vec{x}}
\end{align*}
where the $\pi_i(\vec{x})$ predicates represent constraints obtained in a traditional fashion, while each $\nu_m(\vec{x}_{m,in}, \vec{x}_{m,out})$ is a predicate that: 1) corresponds to a Machine Learning model $m$ from a set $M$; and 2) is satisfied iff the value of the input and output variables match the evaluation of the ML model, i.e.:
$$
\nu_m(\vec{x}_{m,in}, \vec{x}_{m,out}) \Leftrightarrow \vec{x}_{m,out} = m(\vec{x}_{m,in})
$$
The ML components (defining either  constraints or the objective function) are integrated with the rest of the optimization model in a seamless fashion. The emphasis is not on how the ML models are obtained, but on methods for embedding them efficiently and effectively into a combinatorial model. This is the key idea in Empirical Model Learning \cite{LOMBARDI2017343} and can be achieved by either expanding or exploiting the constraint language. We will rely on this distinction to group approaches in this section.

\emph{Embedding a ML model by expanding the language} is a natural solution in the Constraint Programming domain: it requires to introduce a new modeling block (e.g. a new global constraint) and to define an operational semantic (e.g. a propagator). For example, \cite{LOMBARDI2017343} embed a pre-trained Neural Network in CP by associating a ``Neuron Constraint'' to each network unit, and using interval-based reasoning to prune the input/output variables. The approach is extended in \cite{DBLP:journals/constraints/LombardiG16} to two-layer networks via a Lagrangian relaxation. A similar approach is taken in \cite{DBLP:journals/ijait/LallouetL07} and related references by the same authors. In these works, however, the starting point is a collection of examples that implicitly (and approximately) define a constraint. Then, a set of ML classifiers (either Neural Networks or Decision Trees) are learned for checking the consistency of each variable-value pair. Interval based reasoning is then used to generalize the classifiers so that they can work with unbound variables.

The second main method to embed a ML component in a combinatorial model consists in \emph{encoding the ML model using the native language} offered by the optimization technology (e.g. linear constraints and integer variables in Mixed Integer Linear Programming, or linear constraints and boolean predicates in Satisfiability Modulo Linear Real Arithmetic). This encoding is formally a decomposition of the $\nu_m(\vec{x}_{m,in}, \vec{x}_{m,out})$ predicates and should be not only correct, but also effective at supporting the solver at search time.

A simple encoding for Decision (and Regression) Trees in CP and Satisfiability Modulo Theories (LRA in particular) is proposed in \cite{LOMBARDI2017343}, in the context of a thermal-aware workload assignment problem: the encoding is based on modeling each root-to-leaf path as an implication, with the addition of a few redundant constraints. A wider range of CP encodings, based on Multi-Valued Decision Diagrams/\textsc{table} constraints plus discretized numeric attributes, is instead considered in \cite{DBLP:conf/cpaior/BonfiettiLM15}, and benchmarked on the same target problem: these encodings are more complex and computationally expensive, but they also enforce a \emph{much} stronger level of consistency.

A MILP encoding for Decision/Regression Trees (based on associating a binary variable to each root-to-leaf path in the tree) is described in \cite{DBLP:journals/ai/VerwerZY17}, and employed within an auction optimization problem. The encoding uses a small number of integer variable, which is effective at limiting branching, but relies on big-Ms for the linearization of disjunctions, weakening the Linear Programming relaxation.

CP encodings for Random Forests are briefly considered in \cite{DBLP:conf/cpaior/BonfiettiLM15}, although with limited effectiveness. A multi-step HVAC control problem which employs a Deep Neural Network to model the state transition function is considered in \cite{DBLP:conf/ijcai/SayWZS17}: the authors focus on networks based on REctifier Linear Units (ReLUs). For these they provide a MILP encoding strengthened 1) by simple redundant constraints, and 2) by a pre-processing step that sparsifies the network to boost the efficiency of the underlying solver.

\subsection{ML Models as Problem Models}
\label{learningWholeModel}

As an extreme case, we consider works where a ML model (almost) entirely replaces a classical combinatorial optimization model. This happens chiefly when optimization is used for generating counterexamples or for safety verification, and makes such approaches closer to the idea of using optimization to support ML tasks. Formally, approaches in this group deal with problems in the form:
\begin{align*}
  \min z & = x_0 & \mathbf{(P3)} \\
  \text{subject to: } & \nu_m(\vec{x}_{m,in}, \vec{x}_{m,out}) & \forall m \in M \\
  & \vec{x} \in D_{\vec{x}}
\end{align*}
In practice, simple additional constraints on $\vec{x}_{m,in}$ and $\vec{x}_{m,out}$ are usually supported. The main point, however, is that the focus on a specific model structure (e.g. a Neural Network) allows for tailored optimization techniques.

In this context, \cite{FISCHETTICPAIOR2018} model a ReLU based Deep Neural Network using MILP and bound tightening. Starting from a given (correctly classified) example, the approach searches for a minimally-distant input perturbation that invalidates the classification. The method is general, but applied to image classification as a case study. The use of bound tightening proved crucial for the method effectiveness.

In the SMT domain, \cite{DBLP:conf/cav/HuangKWW17} introduce an approach for generating counterexamples for image classification tasks. The authors improve scalability and obtain more meaningful results by replacing basic decisions (i.e. choosing the color of each pixel) with image manipulation operators, applied to an original (correctly classified) example. Conversely, \cite{DBLP:conf/cav/KatzBDJK17} follow a more fundamental, and low-level, approach by explicitly writing a theory solver for Linear Real Arithmetic augmented with ReLUs. The solver attempts to deal with each ReLU in the form $y = \max(0, x)$ by updating $x$ or $y$ to satisfy the relation, and then proceeding with the normal Simplex algorithm. Actual disjunctions are introduced only when a maximum number of updates has been performed on a given ReLU. The method exhibits very good scalability and is successfully  employed to verify properties of an unmanned aircraft control system.

\section{Learning while Solving Problems}
\label{sec:learn_and_solve}

In this section we consider approaches that obtain an approximate model via active learning \emph{while searching for an optimal solution}. Promising candidate solutions are identified and evaluated; then, based on the feedback, the internal model is updated and the whole process repeated. Active learning is also employed in constraint acquisition (e.g. \textsc{QuAcq} and \textsc{Conacq.2}): however, for the approaches in this section, \emph{the main outcome of the process is a solution rather than the learnt model}. Moreover, these approaches focus on learning an explicit \emph{objective function}, rather than an arbitrary model component. Formally, we deal with problems in the form:
\begin{align*}
  \min z & = f(\vec{x}; \vec{w}) & \mathbf{(P4)} \\
  \text{subject to: } & \pi_i(\vec{x}) & \forall i \in I \\
  & \vec{x} \in D_{\vec{x}}
\end{align*}
where $f$ is the objective function, represented using an approximate model with a parameter vector $\vec{w}$ that is learnt at search time. These approaches are employed (e.g.) when dealing with user preferences. In this context, requesting user feedback is referred to as \emph{preference elicitation}, and can involve more than a simple evaluation (e.g. it may require comparisons or explicit changes by the user).

Rather than on preference elicitation in itself (an extensively researched topic), \emph{here we are interested in approaches that use such technique to learn the objective function of a combinatorial problem}. Typically, this is achieved by using preferences to learn a linear combination of ``features'', i.e.:
\begin{equation}
  f(\vec{x}; \vec{w}) = \sum_{j=1}^m w_j \phi_j(\vec{x}) \label{eqn:prefeli}
\end{equation}
where each $\phi_i(\vec{x})$ is a \emph{feature function}. In principle, embedding Equation~\eqref{eqn:prefeli} in combinatorial optimization is easy, as long as the solver can deal with the $\phi_i(\vec{x})$ functions. In practice, identifying a candidate solution requires to estimate both the solution quality \emph{and the degree of uncertainty of the model}: this ensures that promising candidates are not disregarded due to overestimation errors. The need to take into account both these aspects frequently results in restricting assumptions on the supported feature functions.

A method based on SAT Modulo Theories is provided in \cite{Campigotto2011}. The original approach was cast in \cite{DBLP:journals/firai/DragoneTP18} as an instance of a more general framework, grounded also using MILP and two more preference elicitation methods by other authors. The paper assumes discrete or linear numerical features. Other approaches may be viable in principle, but actual application examples to non-trivial combinatorial problems are scarce.


The need to speed-up optimization with expensive objective functions (e.g. defined via numerical simulation) has also motivated the use of \emph{surrogate models in heuristic methods}. In this case an internal, approximate model is employed to reduce the number of needed function evaluations. As an example, \cite{DBLP:conf/aaai/GilanD15} use Gaussian process regression to boost the efficiency of a Genetic Algorithm for sustainable building design.

\section{Related Areas}
\label{relatedAreas}

In this section we review methods that do not strictly fall within the paper focus (i.e. the use of ML to boost modeling in combinatorial optimization), but are nevertheless closely related and likely of interest for the reader.

The iterative refinement loop described in Section~\ref{sec:learn_and_solve} is at the core of \emph{black box optimization approaches based on surrogate models}, recently surveyed in \cite{DBLP:journals/itor/VuDHL17}. Such methods typically tackle problems where the objective function is expensive to compute (e.g. it is defined via a numerical simulator), and use an internal approximate model to reduce the number of evaluations. The models of choice are traditionally (second degree) multivariate polynomials, kriging, or Radial Basis Functions. The solution methods have roots in Mathematical Programming and emphasize dealing with non-linear functions rather than with complex constraints and discrete variables. Surrogate based optimization is \emph{one of the research areas where active learning for optimization has been better investigated}. We have left these approaches out of Section~\ref{sec:learn_and_solve} because their application to combinatorial problems has so far been limited (but the picture is changing rapidly).

The ALAMO system from \cite{DBLP:journals/cce/WilsonS17} is designed to automate the construction of algebraic models of functions that are expensive to evaluate. The approach relies on active learning, and generates new sampling points by maximizing the estimated error of the current model. It can be considered an extreme form of surrogate based optimization where minimizing the error is the only objective. It was originally designed to learn objective/constraint functions for numeric (rather than combinatorial) optimization problems, and for this reason has been left out of Section~\ref{sec:constr_acq}.






OptQuest \cite{OptQuest} interleaves simulation for evaluating the problem objective and optimization (based on scatter search). OptQuest uses predictive models within the search engine to establish promising research directions. These predictors can be either based on multivatiate linear regression or neural networks that are trained during search. The approach has not been considered in the Section~\ref{sec:learn_and_solve} because the ML component is not strictly used for modeling.

In Control Theory, similar concepts have been deeply studied in the context of Model Predictive Control \cite{DBLP:journals/cce/ChristofidesSPL13} (MPC). MPC chooses control actions by repeatedly solving an online constrained optimization problem, which aims at minimizing a performance index over a finite horizon based on predictions obtained by a system model. The model is obtained through a system identification methodology that is capable to accurately predict the system dynamics. System identification starts from a data set, a set of candidate models (the model structure) and a rule by which candidate models can be assessed against data (e.g. the least square selection rule). The selected model is then validated and refined. Despite some similarities between system identification and ML and between MPC and optimization exist, these methods are inherently different in the decisions they take. Control models act on-line by taking and applying decisions to the system. Combinatorial optimization is in general concerned with more strategic decisions that have a longer time horizon and are not applied at optimization time.

\section{Common Themes and Shared Issues}
\label{Transversal_issues}

In this section we discuss some themes and issues that tend to be shared by all approaches attempting to modeling activities via ML. Some of them have been recognized and tackled (at least in certain subfields), while others have been generally neglected or only recently discovered. Some themes have been briefly discussed in the previous sections, but it is worth to treat them here in a more systematic fashion.

\mypar{Dealing with the Model (In)accuracy:} When ML makes up part of the problem model, we explicitly acknowledge that the solutions may suffer from approximation errors. This is exacerbated by the fact that \emph{the solutions (or more accurately the ML input configurations) visited at search time may be considerably different from those considered at training time}. There are two main ways to deal with this situation.


First, for approaches that include a passive learning stage, the problem can be mitigated by ensuring that the training set provides a sufficiently uniform coverage of the search space. This can be done by choosing the training data according to Design of Experiments principles (e.g. via Latin hypercube sampling). This approach has been well investigated in surrogate-based optimization, because the availability of a simulator (or the chance to perform physical experiments) provides the opportunity to design an ad-hoc training set.

The second (more rigorous) approach consists in using active learning. The simple approach of evaluating a solution and updating the ML model is sufficient to guarantee convergence, provided that the updates improve the model accuracy in a roughly monotonic fashion. However, the simulators frequently employed in active learning may introduce additional approximation. Finally, both the first and the second method are not easily applicable to collections of historical data.

\mypar{Optimizer Bias:} In the case of optimization (say, minimization) problems, the solver will naturally be attracted by solutions with a low objective value and will actively avoid solutions with large objective value. In practice, \emph{such values may have more to do with approximation errors than with the actual quality of the solution}. In case of an underestimated objective, active learning is sufficient to ensure convergence to high quality solutions in most cases. Dealing with overestimation, however, additionally requires access to information about the model accuracy.

Some techniques (e.g. Gaussian processes, Support Vector Machines) provide this information in a rigorous fashion under specific conditions (e.g. smoothness). For this reason they have been thoroughly investigated in the context of black-box optimization and preference elicitation. For example, assuming that the ML model provides a confidence interval for the problem objective, the solver may be instructed to focus on the lower-end of such interval rather than on the actual prediction. More in general, a solution may be accepted as long as it is \emph{sufficiently likely} to improve over the best known value. Finding effective methods to combine uncertainty and solution quality information is however far from trivial, and the details of this integration are the area where multiple black-box and preference elicitation methods tend to differ the most.

Information about the model uncertainty is in fact available for many ML methods: (Deep) Neural Network classifiers have softmax scores, Decision Trees report misclassified examples on their leaves, Regression Trees can do the same with output variance, and Random Forests provide vote counts. To the best of our knowledge, however, this kind of information has never been thoroughly exploited in optimization.

\mypar{Passive vs Active Learning:} In general, \emph{active learning can provide significant advantages over passive learning} in terms of solution accuracy and convergence. \emph{However, the technique is not without disadvantages}. In first place, active learning requires the ability to evaluate solutions, which is not available for example when working with historical data. Second, training at search time may be prohibitively expensive, depending on the response times that are needed for the considered optimization application. The cost may stem from the time for a function evaluation and from the number of examples needed to obtain a reasonably accurate model.

The last point makes it particularly difficult to combine active learning with data hungry ML models, such as Deep Neural Networks.
Due to the effectiveness of DNNs and the practical benefits of active learning, identifying techniques to combine the two is a promising research topic. Intuitively, it would require finding methods for making significant updates to a pre-trained DNN using only a small number of examples.

\mypar{Deterministic vs Non-Deterministic Systems:} Machine Learning is frequently used to model systems that are not deterministic, e.g. that are stochastic or simply exhibit an inherently degree of inconsistency (e.g. user evaluations). In practice, \emph{in both cases the same input example may lead to different results, with multiple evaluations following some kind of probability distribution}.

Many of the methods described in this work (e.g. those in Section~\ref{learningConstraints} and classical black-box optimization) tend to overlook this behavior. In the case of passive learning with historical data, the training test implicitly encodes information about the probability distribution, which ends up mitigating the problem. When the training set is crafted via Design of Experiment, or when active learning is used, extra care should be put when the underlying system is non-deterministic: for example, one could resort to multiple evaluations of the same (or similar) examples, or make use of ML models that are robust w.r.t. inconsistent evaluations. The latter approach is typically employed in preference elicitation.

A more subtle issue is that \emph{a model trained for maximum likelihood over a stochastic system may lead to sub-par results when used in combinatorial optimization}. This fact has been recently recognized in \cite{NIPS2017_7132}, and tackled via so called end-to-end task based learning, i.e. by explicitly taking into account the optimizer during passive training. Currently, the approach is viable only for a specific class of (convex) optimization problems, and a straightforward generalization may be very computationally expensive.

More in general, accuracy issues are not the only reason why updating the model may become necessary. The problem specification itself may be incomplete, or change over time. This is explicitly  recognized in  \cite{DBLP:journals/expert/BessiereRGKNNOP17}, which proposes the Inductive Constraint Programming Loop as a general framework do deal with model updates.

\mypar{Accuracy vs Optimization:} In our opinion, however, the most relevant and less investigated issues relate to recognizing the approximate nature of the model. In first place, \emph{when dealing with an approximate model, the practical value of finding global optima appears questionable}. This does not hold for approaches that attempt to verify properties of ML models, and holds to a lesser degree for problems where an accurate model is available, but expensive to compute. In all other cases, reaching global optimality may even be detrimental, if the corresponding solution happens to be poor in terms of real world robustness. Intuitively, it should be possible to exploit uncertainty information to stop the search process earlier, depending on the level of accuracy of the model. However, no such attempt has been considered in the literature to the best of our knowledge.

Finally, when the ML model and the combinatorial model need to be co-designed, there exists an interesting \emph{trade-off between model accuracy and suitability for optimization}. Namely, a complex model (e.g. a deeper network) may be more accurate, but more difficult to optimize, possibly leading to worse results in practice. To the best of our knowledge, this kind of trade-off has never been thoroughly investigated.

\section{The Use of ML beyond Modelling}
\label{general}
Beside the modeling activity, Machine Learning has been exploited also in the solution of the problem by improving search. For instance in Large Neighborhood search an initial solution is gradually improved by alternately destroying and repairing the solution. The destroying part is performed by selecting and fixing some fragment of the solution and the repairing part is based on search. Machine Learning, and in particular, reinforcement learning has been used in  \cite{Mairy_reinforcedadaptive}  to tune the search limit, the fragment size and the fragment selection procedure.

In more traditional tree search, Machine Learning has been used either to select the best heuristics \cite{DASH} from a portfolio, or to guide search by estimating the best variable-value selections and/or a bound on the objective: the estimate may come from a model trained off-line over available solutions \cite{DeepLearningHeuristics,DNN-CPAIOR2018}, or from sampling (e.g. \cite{DBLP:conf/cp/LothSHS13}). Both methods can be effective, but sampling may incur significant overhead at solution time, while training off-line may lead to a lack of generality: if the Machine Learning model is trained on a given problem dimension, then it can be used to guide the search in instances of lower or equal dimension, but it does not scale to larger dimension problems. This difficulty is partially overcome in \cite{DBLP:journals/corr/BelloPLNB16} by using pointer networks and reinforcement learning, but the technique requires to engineer the network structure for the considered problem, limiting in part its flexibility. In general, the way to generalise these approaches and improve their efficiency is a very intriguing direction for future research.


Machine Learning has also been applied successfully in automatic algorithm portfolio selection. This line of research has received more attention w.r.t. the more recent topics described above.
An algorithm portfolio consists of a set of algorithms, and portfolio selection is the problem of choosing
the best one for an input instance. A highly successful approach in satisfiability (SAT) is SATzilla \cite{DBLP:journals/jair/XuHHL08}, other known systems are Hydra, CP-Hydra, and SMAC. It is based on the identification of features that characterize problem instances and are related to instance hardness. In general this cannot be done automatically, but rather by capturing the knowledge of a domain expert. Among all approaches developed for this purpose, SMAC stands out for its use of active learning and an approximate model trained at search time (see \cite{DBLP:conf/lion/HutterHL11}). Other works have focused instead on using ML to choose the best technique to solve a sub-problem, and specifically to choose the best propagator for a given constraint in CP (e.g. \cite{DBLP:conf/ijcai/BalafrejBP15}).


ML techniques are also applicable in other contexts like performance predictions (see e.g. \cite{DBLP:journals/ai/HutterXHL14}, \cite{DBLP:conf/cpaior/MalitskyS12}) to devise a schedule with time allocations for each algorithm in the portfolio, which can then be applied sequentially or in parallel.

%

\section{Conclusions}
\label{sec:conclusions}

The hybridization of ML and optimization is a broad research field that has been gaining considerable popularity in recent years. This topic has been so far tackled in relative isolation in multiple domains, somehow hindering the research progress. In this work, we have provided a first cross-disciplinary survey focused on \emph{the use ML to boost the modeling activity  of combinatorial problems}. We have attempted a classification and highlighted closely related fields. More importantly, we have identified a number of common themes and issues that have been addressed in different context, which may serve as a guide to promote cross-fertilization efforts.

\bibliographystyle{named}
\bibliography{ijcai18}

\end{document}